\documentclass[a4paper]{article}

\usepackage{INTERSPEECH2022}

\usepackage{float}

\usepackage[all]{nowidow}

\newcommand{\comment}[1]{}

\usepackage{caption}
\usepackage{subcaption}

\usepackage{siunitx}

\usepackage{booktabs}
\usepackage{tabularx}

\usepackage{url}

\usepackage{color}
\definecolor{deepblue}{rgb}{0,0,0.5}
\definecolor{deepred}{rgb}{0.6,0,0}
\definecolor{deepgreen}{rgb}{0,0.5,0}

\usepackage{listings}
\usepackage{etoolbox}
\newcounter{BalanceAtReference}
\newcounter{ReferenceIndexForBalancing}
\makeatletter
\def\@balancelastpageonce{%
	\ifnum\value{ReferenceIndexForBalancing}=\value{BalanceAtReference}
	\newpage
	\else
	\relax
	\fi
	\stepcounter{ReferenceIndexForBalancing}
}
\pretocmd{\bibitem}{\@balancelastpageonce}
{} 
{\@latex@error{Patching \bibitem failed}{\@ehd}}
\makeatother

\title{Finstreder: Simple and fast Spoken Language Understanding \\ with Finite State Transducers using modern Speech-to-Text models}
\name{Daniel Bermuth, Alexander Poeppel, Wolfgang Reif}
\address{University of Augsburg, Institute for Software \& Systems Engineering}
\email{\{daniel.bermuth,alexander.poeppel,reif\}@informatik.uni-augsburg.de}

\begin{document}

\maketitle
  
\begin{abstract}
	In \textit{Spoken Language Understanding}~(SLU) the task is to extract important information from audio commands, like the intent of what a user wants the system to do and special entities like locations or numbers.
	This paper presents a simple method for embedding intents and entities into \mbox{\textit{Finite State Transducers}}, and, in combination with a pretrained general-purpose \textit{Speech-to-Text} model, allows building \mbox{SLU-models} without any additional training. Building those models is very fast and only takes a few seconds. It is also completely language independent.
	With a comparison on different benchmarks it is shown that this method can outperform multiple other, more resource demanding SLU approaches.
\end{abstract}

\noindent\textbf{Index Terms}: spoken language understanding, speech to intent, offline voice assistant, finite state transducer decoding

\section{Introduction}
\label{sec:intro}

When building models for \textit{Spoken Language Understanding}~(SLU), there are two alternative approaches: Using two separate stages of transcribing the spoken command to text (\textit{Speech-to-Text}, STT) and then extracting the useful information out of the transcribed sentence (\textit{Natural Language Understanding}, NLU), or using a direct SLU approach which combines the two parts into one single model. The first has the benefit, that the two models can be trained independently and the STT-model can often be used across multiple different domains, while the NLU-model can be trained relatively quickly. The second approach on the other hand is often more accurate, because it does not have the problem that errors from the STT transcription are propagated into the NLU module.

\vspace{9pt}
Recent systems for NLU or SLU parsing usually build upon neural networks as feature extractors. While they generally achieve a high recognition accuracy, the downside is that training those networks can take a lot of time. In this work a completely different approach is investigated, which does not require any special training, and therefore allows very fast creation of SLU models. It uses \textit{Finite State Transducers}~(FSTs) instead of neural networks for SLU parsing.  The speech recognition part still uses a neural network, but this only has to be trained once per language on general-purpose STT tasks. In this work a \textit{Quartznet} \cite{QNET} model from previous work in \textit{\mbox{Scribosermo}}~\cite{SCRSO}, as well as a \textit{Conformer} \cite{CNET} model from \textit{NeMo}~\cite{NEMO}, converted to \textit{tensorflow-lite} with \textit{Scribosermo}, are used. Both models, after conversion to \textit{tflite} and quantization, can run faster than real-time on a RaspberryPi4. 

A comparison on multiple benchmarks shows that the performance of this approach is highly competitive with other solutions, despite the fact that the concept is very simple and the models are built in a few seconds. 

The source-code of the presented method for training-free SLU parsing, named \mbox{\textit{finstreder}}, as well as the models from \mbox{\textit{Scribosermo}}, can be found at: \mbox{https://gitlab.com/Jaco-Assistant}

\vspace{9pt}
Using FSTs in speech recognition tasks is not a new idea and was quite common some years ago \cite{WFSTSR, SRWFST}. FSTs are used by \textit{Kaldi}~\cite{KALDI}, where a neural network outputs phonemes which are then decoded to sentences with a combination of multiple FSTs. 
Some works already explored the usage of FSTs for parsing NLU information by adding semantic tags into the FSTs, either from textual inputs~\cite{GDASLU, HRBRNN} or from speech transcription hypotheses of a hidden Markov model~\cite{FSTSI}. The authors of~\cite{ISLUWC} explored how to use grammar fragments with embedded tags to improve word confusions. In \cite{RAWDS} a dialog system is described which transforms textual user utterances into response sentences using 
weighted FSTs, with the goal to be able to run a full back-and-forth dialog with the users. It was extended by~\cite{EXWDS} to accept n-best hypotheses from a tri-phone model acoustic model, which were combined with an additional 3-gram language model, as input.
\textit{Eesen}~\cite{EESEN} introduced FST-decoding to models outputting character-based \textit{Connectionist Temporal Classification}~(CTC)~\cite{CTCL} labels, similar to the \textit{Quartznet} model of \textit{Scribosermo}. \textit{Alexa} also uses FSTs for its skill kit, but keeps separate models for STT and NLU~\cite{JSTASK}. This work follows a very similar decoding approach as \textit{Eesen}, which allows using recent CTC-based STT models (in difference to~\cite{FSTSI,ISLUWC,EXWDS}), but alters the \textit{Grammar-FST} (explained in the next chapters) to embed NLU information into it, similar to the semantic tagging of~\cite{HRBRNN,FSTSI,ISLUWC}, which allows combining the two distinct STT+NLU models into a single SLU decoder. 
\mbox{\textit{OpenFST}}~\cite{OFST} is used as library for handling the FSTs.

\section{Foundations}
\label{sec:foundat}

\begin{figure}[!htbp]
	\centering
	\includegraphics[width=0.7\linewidth]{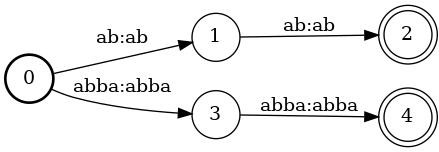}
	\caption{Simple Grammar-FST which accepts the sentences \textit{``ab~ab"} or \textit{``abba~abba"}. The part in the transition on the left side of the colon denotes the input which is required to make the transition and the one on the right side the output which is received afterwards.}
	\label{fig:sg_grammar}
\end{figure}

In general, the language models used for decoding speech features with FSTs can be split into different parts: 

\noindent(1) A \textit{Grammar-FST}, conventionally denoted as \textit{G}, which stores the information of complete sentences. Figure~\ref{fig:sg_grammar} shows a very simple grammar that can accept exactly two different sentences.

\begin{figure}[!htbp]
	\centering
	\includegraphics[width=\linewidth]{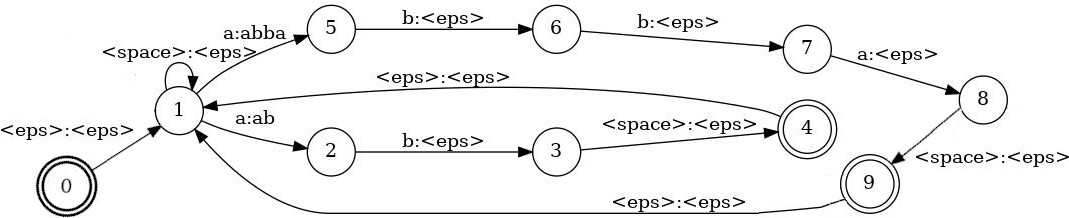}
	\caption{The Lexicon-FST for the words \textit{``ab"} and \textit{``abba"}.}
	\label{fig:sg_lexicon}
\end{figure}

\noindent(2) A \textit{Lexicon-FST}, denoted as \textit{L}, which builds words out of characters, as shown in Figure~\ref{fig:sg_lexicon}. For later optimization of the FST, a disambiguation symbol is required, which ensures that the path for \textit{``ab"} doesn't also accept \textit{``abab"}. Instead of introducing a special symbol, each word has to end with a space. 

\begin{figure}[!htbp]
	\centering
	\includegraphics[width=\linewidth]{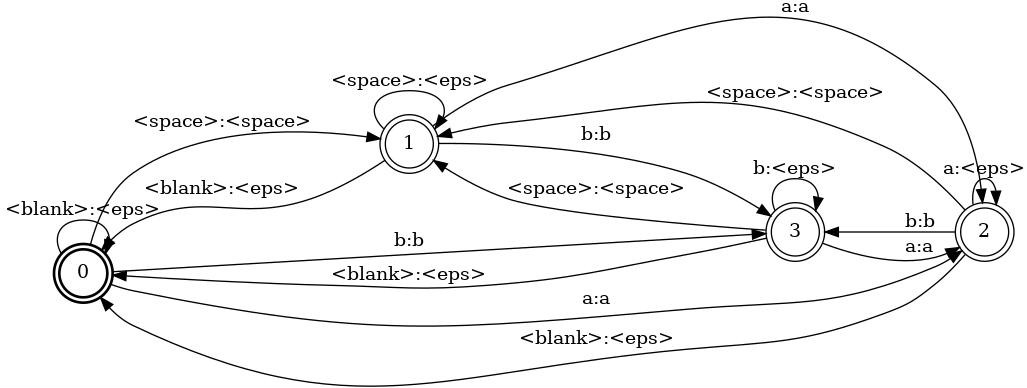}
	\caption{The Token-FST which merges CTC-labels like  \textit{``aaab~ab-b"} to the characters \textit{``ab~abb"}.}
	\label{fig:sg_token}
\end{figure}

\noindent (3) In the case of CTC-Inputs, a \textit{Token-FST (T)} which converts the frame-level CTC-labels to characters. CTC-labels are commonly used in STT-systems and contain a special \textit{\textless{}blank\textgreater{}} label (here also written as \textit{``-"}) besides the default alphabet characters. To generate normal text from those labels, all repeated tokens are merged into one single character, except they are separated by a \textit{\textless{}blank\textgreater{}} label, which is removed after the merging step. Figure~\ref{fig:sg_token} shows a simple FST for the alphabet \textit{``\textless{}space\textgreater{},a,b,\textless{}blank\textgreater{}"}. To allow the usage of \textit{sentencepiece}-style labels \cite{SNTPC} from the \textit{Conformer} model, this  FST was extended in a way that the pieces are split into single characters.

\begin{figure*}[!htbp]
	\centering
	\includegraphics[width=\linewidth]{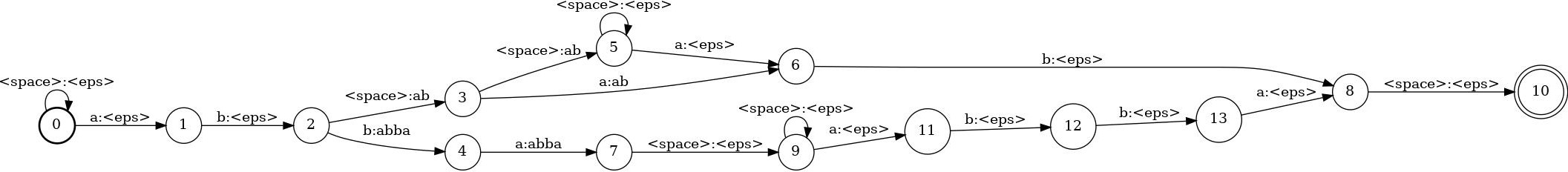}
	\caption{Optimized LG-FST which has characters as input and the sentences from the simple grammar above as output.}
	\label{fig:sg_lgopt}
\end{figure*}

Instead of using three single FSTs while decoding, they can be \textit{composed} together. Afterwards an optimization step can be applied, which restructures the graph to remove unnecessary or duplicated transitions. Figure~\ref{fig:sg_lgopt} shows the composition of the Lexicon-FST with Grammar-FST. The result can then be composed with the Token-FST, similar to the approach in \textit{Eesen}~\cite{EESEN}, so that the composition can handle CTC-labels as input.

\begin{figure}[H]
	\centering
	\includegraphics[width=\linewidth]{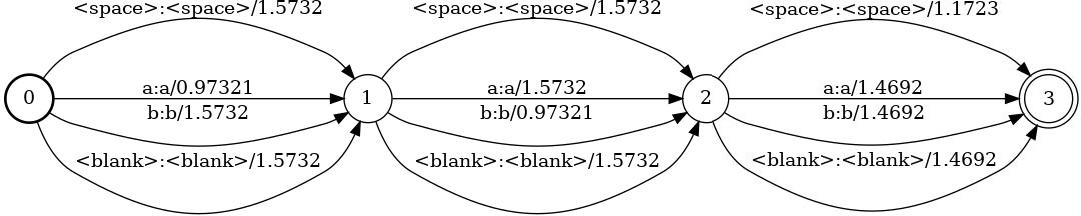}
	\caption{An Input-FST with the most probable path \textit{``ab"}, ending with an extra space.}
	\label{fig:sg_input1}
\end{figure}

\vspace{9pt}
The input of the combined \textit{TLG} model also has to be in form of a FST. Thus the step-by-step labels are converted into a state-by-state FST with one transition for each character between the two states of a timestep (Figure~\ref{fig:sg_input1}). The label probabilities in the range $[0\text{-}1]$, where higher is better, have to be converted so that a lower value is better. The normalization approach of~\cite{GSAMP} is used for this. Additionally, an extra timestep is added as last timestep, which has the space character as its highest probability, to ensure that the last word ends with a space (the disambiguation symbol of the lexicon).

\begin{figure}[H]
	\centering
	\includegraphics[width=0.5\linewidth]{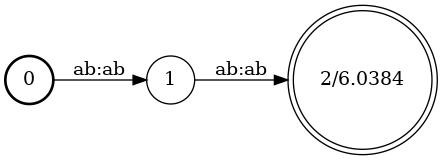}
	\caption{The Output-FST which accepts the sentence \textit{``ab~ab"} as most probable (shortest) path, with the total path length in the last state.}
	\label{fig:sg_output1}
\end{figure}

The reason behind the probability conversion is that after composition, the resulting FST includes all possible (acceptable) combinations of words that can be created with the given input. Since only the best matching result is desired, the shortest path algorithm from \textit{OpenFST} can be used to search the path with the lowest transition weights (which came from the CTC-labels). The result is a new FST, which accepts the best matching input words (Figure~\ref{fig:sg_output1}). The decoding algorithm can then traverse this FST state-by-state and extract the output of each transition, which can then be merged into a returnable sentence.

\section{Finstreder's approach}
\label{sec:sluap}

In \textit{Spoken Language Understanding}, the goal is to extract important features from a spoken utterance. In the case of users speaking voice commands, which is a common area for such applications, the software developers are mainly interested in the \textit{intent}, the main goal of the command, and in special \mbox{\textit{entities/slots}}, like names, locations or numbers. 

This chapter shows how intent and entity information can be included directly in the Grammar-FST, with only slight adjustments. The decoder is then able to transcribe the text and extract the required information in one single step, thus removing the time-consuming step of training a specialized NLU-model.

\subsection{Building the models}

Building the new Grammar-FST is separated into four different steps. As input a \textit{json}-file following the syntax structure of \textit{Jaco}~\cite{JACO} is required:

\noindent
\begin{minipage}{\linewidth}
\begin{lstlisting}
{
  "intents": {
    "get-looks": [
      "(is a|are) [---](animal) cute"
    ]
  },
  "lookups": {
    "animal": [
      "whitemargin stargazer",
      "atlantic stargazer",
      "aye aye",
      "(hairy frogfish)->striated frogfish"
    ]
  }
}
\end{lstlisting}
\end{minipage}

\begin{figure}[H]
	\centering
	\includegraphics[width=\linewidth]{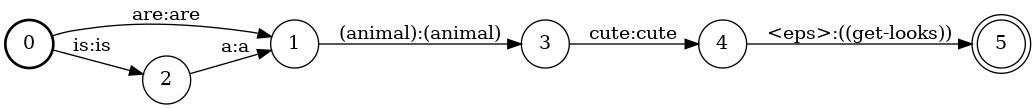}
	\caption{Example Intent-FST with placeholder for entity values and the intent name at the end.}
	\label{fig:itc_intent}
\end{figure}

\vspace{9pt}
In the first step, a new text file for each intent is created, including all example sentences. The files are then converted to FSTs, either using \textit{n-grams} or as \textit{fixed} grammar. All final states are then forwarded using an \textit{epsilon} transition to include the intent name directly into the FST, as shown in Figure~\ref{fig:itc_intent}.

\begin{figure}[H]
	\centering
	\includegraphics[width=\linewidth]{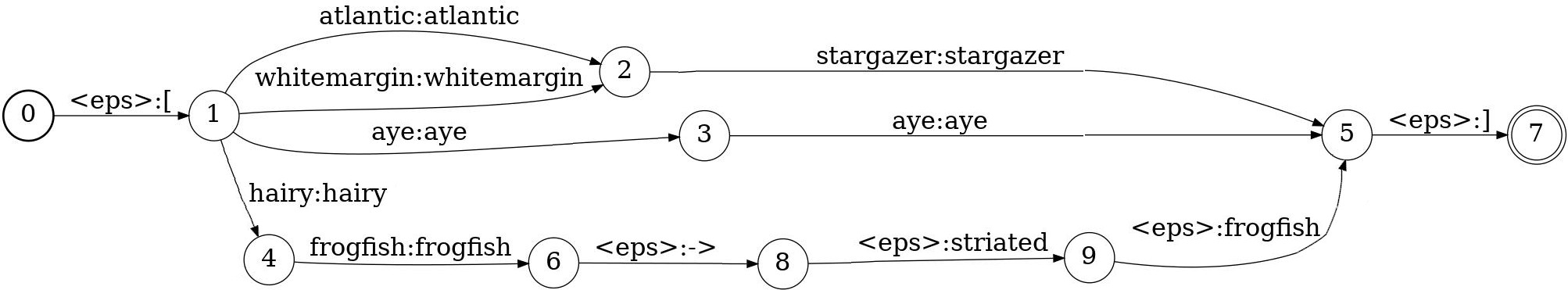}
	\caption{Example Slot-FST with special markers and synonym replacements (for the \textit{frogfish}).}
	\label{fig:itc_animal}
\end{figure}

In the next step one FST for each slot is created. It also includes special symbols as slot markers and handles synonym replacements, by adding the synonym value after the actual value into the graph. See  Figure~\ref{fig:itc_animal}.

In the third step, the Slot-FSTs are inserted directly before their placeholders in the Intent-FSTs.
The placeholder itself is kept, because it is later needed to determine the corresponding slot type.
At last the merged Intent-FSTs are each composed with the Lexicon-FST.

\comment{
\begin{figure*}[!htbp]
	\centering
	\includegraphics[width=\linewidth]{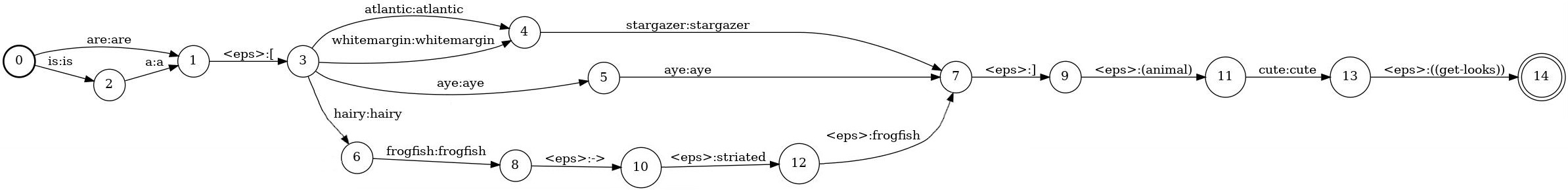}
	\caption{Combined Intent-Slot-FST.}
	\label{fig:itc_merged}
\end{figure*}
}

\subsection{Decoding}

Decoding works as follows: First an Input-FST is built using the CTC-labels. Then it is composed with the Token-FST. Afterwards the resulting IT-FST is composed with each LG-FST of the different intents. The shortest path in each resulting FST is calculated and the overall shortest path is returned as result.

To improve and speed up decoding, several optimizations were implemented. Instead of uniting the LG-FSTs for each intent and composing it with the Token-FST into a single large TLG-FST, they are kept separate. Using multiple FSTs for the intents allows automatic parallelization with the IT-FST if there are many intents. This also allows including or excluding specific intents for each request.
To exclude less necessary input characters, one parameter allows removing labels where the CTC-probability is below the \textit{top-k} best values of each timestep. Another one removes all labels where the probability is below the mean value of the k-most-probable character over all timesteps (\textit{mean-k}). Both parameters can greatly speed up decoding time with a very small impact on accuracy. A well working combination is $\textit{top-k}=5\text{-}12$ and $\textit{mean-k}=21$.
For reweighing the input labels, a parameter was introduced that executes exponential scaling of the probabilities in a timestep, which increases or decreases the relative difference between label probabilities. A second parameter can be used for linear scaling of the input weights versus the grammar weights from the \textit{ngram} model.

\section{Experiments}
\label{sec:exper}

The performance of \textit{finstreder} was evaluated in multiple benchmarks. The first three benchmarks were reused from previous work in the \textit{Jaco} project \cite{JACO}. 
Since the \textit{Quartznet} model was already used in \textit{Jaco}, combined with NLU models trained with \textit{Rasa}~\cite{RASA}, the experiments with it allow a direct comparison between the two semantic parsing methods. The STT models \mbox{(\textit{QuartzNet-15x5}$/$\textit{ConformerCTC-L})} reach a greedy \textit{Word Error Rate} (WER) of \SI{4.6/2.2}{\percent} on \textit{LibriSpeech}~\cite{LIBSPE} in English and \SI{17.5/8.0}{\percent} on \textit{CommonVoice}~\cite{COMV} in French. 
All of the benchmark code is released in the same repository as this work.

\subsection{Spoken Language Understanding}

\vspace{9pt}
The \textit{Barrista} benchmark was published by \textit{Picovoice} \cite{RHIBEN} and consists of 620 commands of different people ordering coffee in English. The audio is mixed with different volume levels of background noise from cafe and kitchen environments. An example of a command would be: \textit{``i'd like a [medium roast] [large] [mocha] with [lots of cream] and [a little bit of brown sugar]"}. A command is correctly detected if the intent, as well as all the slots, could be retrieved by the assistant. This metric will be used for the following benchmarks as well.

\begin{figure}[!htbp]
	\centering
	\includegraphics[width=\linewidth]{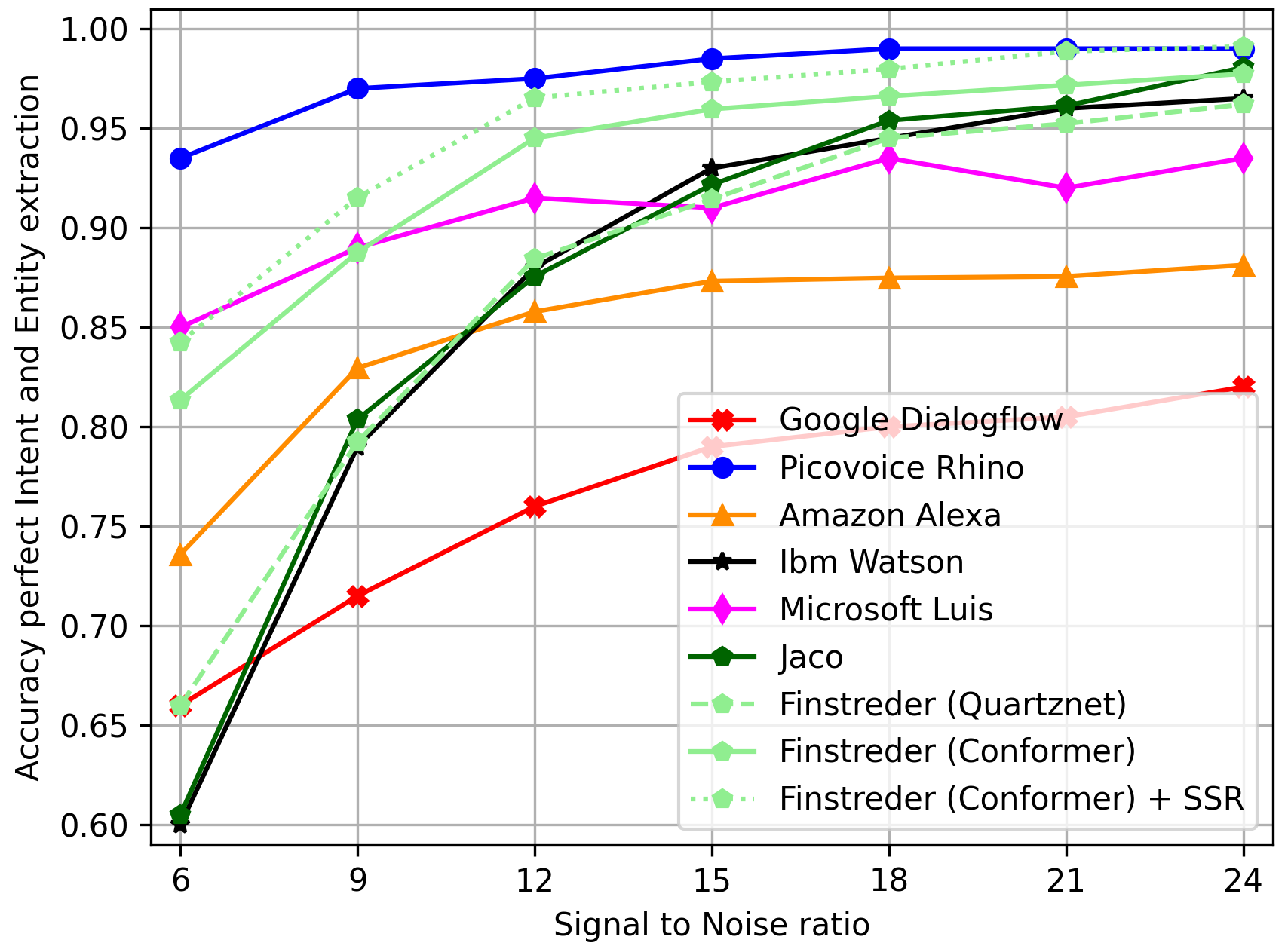}
	\vspace{-7pt}
	\caption{Benchmark coffee orders with noisy backgrounds. The results of \textit{DialogFlow}, \textit{Watson}, \textit{Luis} and \textit{Rhino} have been taken from \cite{RHIBEN}, the results of \textit{Alexa} and \textit{Jaco} from \cite{JACO}.}
	\label{fig:bb}
\end{figure}

The results in Figure~\ref{fig:bb} show that \textit{finstreder} with the \textit{Quartznet} model performs as well as \textit{Jaco} and \textit{Watson}, and, like \textit{Jaco}, has some problems with noisy backgrounds, which is most likely related to poorer recognition in the acoustic model. In comparison to that, using the \textit{Conformer} model improves the results, especially in noisy environments.

Because the SLU model is built only from text files, it is very simple to add frequent transcription errors as synonyms into the dialog definition. After adding similar sounding synonyms as replacements (+SSR) for the three most common errors (\textit{(rose$\vert$roast)}, \textit{(ons$\vert$ounce)}, \textit{(ice~moka$\vert$iced~mocha)}), and rebuilding the SLU model, the accuracy improved notably.

\vspace{9pt}
The \textit{SmartLights} benchmark from \textit{Snips} \cite{SNSLU} tests the capability of controlling lights in different rooms. It consists of 1660 requests which are split into five partitions for a 5-fold evaluation. A sample command could be: \textit{``please change the [bedroom] lights to [red]"} or \textit{``i'd like the [living room] lights to be at [twelve] percent"}. The benchmark results are presented in Table~\ref{tab:ben_slc}. The performance of \textit{finstreder} with the \textit{Quartznet} model is on-par with the two-step approach of \textit{Jaco} in this benchmark, as well as with the E2E-SLU approach of \textit{AT-AT} \cite{ATAT}, and outperforms them with the \textit{Conformer} model. \textit{Snips}~\cite{SNIPS}, which was a voice assistant, but is not available anymore, uses \textit{Kaldi} as STT module and their own NLU module. The models were trained on online servers and could then be downloaded, which enabled the assistant to decode the voice requests without an internet connection. \textit{Lugosch et al.}~\cite{SYNSLU} use the features of a pretrained STT-network and add a SLU-decoder on top of it. Their model is then finetuned with a combination of real and synthetic speech data. \textit{AT-AT}~\cite{ATAT} uses two single encoder networks for audio and text inputs, but combines them in a single decoder network, with the advantage that the decoder can be finetuned with text only data and the network can also apply the learned knowledge to audio inputs. 

\begin{table}[!htbp]
	\footnotesize
	\caption{Results on \textit{SmartLights} dataset.}
	\vspace{-5pt}
	\label{tab:ben_slc}
	\centering
	\begin{tabular}{lcc}
		\toprule
		\textbf{} & \textbf{Accuracy} & \textbf{WER} \\
		\midrule
		\textit{Google} \cite{SNSLU} & $0.793$ & $-$ \\
		\textit{Snips} \cite{SNSLU} & $0.842$ & $-$ \\
		\textit{Alexa} \cite{JACO} & $0.792$ & $-$ \\
		\textit{Houndify} \cite{JACO} & $0.545$ & $0.108$ \\
		\textit{Jaco} \cite{JACO} & $0.854$ & $0.108$ \\
		\textit{Finstreder (Quartznet)} & $0.848$ & $0.107$ \\
		\textit{Finstreder (Conformer)} & $\textbf{0.880}$ & $\textbf{0.061}$ \\
		\textit{AT-AT} \cite{ATAT} & $0.849$ & $-$ \\
		\textit{Lugosch et al.} \cite{SYNSLU} & $0.714$ & $-$ \\
		\bottomrule
	\end{tabular}
\end{table}

\vspace{9pt}
The \textit{SmartSpeaker} benchmark tests the performance of reacting to music player commands in English as well as in French. The benchmark is from \textit{Snips} \cite{SNSLU}, too, and is the only one that could be found which includes a language other than English. It has the difficulty of containing many artist or music tracks with uncommon names in the commands, like \textit{``play music by [a boogie wit da hoodie]"} or \textit{``I'd like to listen to [Kinokoteikoku]"}. As shown in Table~\ref{tab:ben_ssc}, using \textit{finstreder's} SLU approach greatly improves accuracy compared to \textit{Jaco's} STT+NLU concept. This could partially be explained by a different handling of the artist names in the \textit{ngram} language models. The \textit{ngrams} of \textit{Jaco} directly include the names, which allows the parser to leave out some parts of the names or mix them up, whereas the ones of \textit{finstreder} only have a placeholder which then has to be matched exactly.

\begin{table}[htb]
	\footnotesize
	\caption{Accurracy on \textit{SmartSpeaker} dataset.}
	\vspace{-5pt}
	\label{tab:ben_ssc}
	\centering
	\begin{tabular}{lcc}
		\toprule
		\textbf{} & \textbf{English} & \textbf{French} \\
		\midrule
		\textit{Snips}~\cite{SNSLU} & $0.687$ & $0.751$ \\
		\textit{Google}~\cite{SNSLU} & $0.478$ & $0.423$ \\
		\textit{Jaco}~\cite{JACO} & $0.627$ & $0.480$ \\
		\textit{Alexa}~\cite{JACO} & $0.455$ & $\textbf{0.889}$ \\
		\textit{Finstreder (Quartznet)} & $0.776$ & $0.778$ \\
		\textit{Finstreder (Conformer)} & $\textbf{0.804}$ & $0.783$ \\
		\bottomrule
	\end{tabular}
\end{table}

\vspace{9pt}
In the \textit{TimersAndSuch} benchmark \cite{TIASU} common use-cases involving numbers are tested. It includes commands like  \textit{``set an alarm for 9:24 a.m."} or \textit{``compute 12.15 plus 26.9"}. The main difficulty is to recognize many different numbers, there are only very few command prefixes (like \textit{``set an alarm for"},~...), therefore a \textit{fixed} grammar model is used instead of a \textit{2-gram} model, which showed a better performance in the other benchmarks. The results in Table~\ref{tab:ben_tas} show that \textit{finstreder} performs generally well in this task too, and could outperform the benchmark's large baseline SLU model, which was trained specifically for this task, as well as the model from \cite{TASSB}, which used the baseline's architecture but included additional unsupervised training for the model's encoder. A test with \textit{Alexa} was skipped, because \textit{Alexa's} built-in number entity did not understand numbers with decimals.

\begin{table}[!htbp]
	\footnotesize
	\caption{Accuracy on \textit{TaS} dataset.}
	\vspace{-5pt}
	\label{tab:ben_tas}
	\centering
	\begin{tabular}{lc}
		\toprule
		\textit{TaS-baseline} \cite{TIASU} & $0.816$ \\
		\textit{SpeechBrain} \cite{TASSB} & $0.940$ \\
		\textit{Finstreder (Quartznet)} & $0.900$ \\
		\textit{Finstreder (Conformer)} & $\textbf{0.954}$ \\
		\bottomrule
	\end{tabular}
\end{table}

\vspace{9pt}
\textit{FluentSpeechCommands} \cite{FSC} tests simple voice assistant requests. It includes commands like  \textit{``turn up the [bathroom] temperature"},  \textit{``switch the lights on"} or \textit{``go get me my [shoes]"}. The benchmark is run with a \textit{fixed} grammar model, too. The results can be found in Table~\ref{tab:ben_fsc}.
\textit{Kim et al.}~\cite{TKNDIS} are combining a textual \textit{BERT} model with a \textit{vq-wav2vec-BERT} model and a \textit{DeepSpeech2} acoustic model to a large SLU end-to-end network. This followed the idea of knowledge distillation from the text model to the speech encoder during training.

Even though the focus of this work is on training-free SLU, it is of course possible to finetune the acoustic model with \textit{Scribosermo} on the used dataset. After a short training (about $1$:$20$\,h on a single RTX2070) the model (\textit{+AMT}) reaches state-of-the-art performance.

\begin{table}[!htbp]
	\footnotesize
	\caption{Accuracy on \textit{FSC} dataset.}
	\vspace{-5pt}
	\label{tab:ben_fsc}
	\centering
	\begin{tabular}{lc}
		\toprule
		\textit{Alexa} & $0.987$ \\
		\textit{FSC-baseline} \cite{TIASU} & $0.988$ \\
		\textit{Cao et al.} \cite{SEISL} & $0.990$ \\
		\textit{FANS} \cite{FANS} & $0.990$ \\
		\textit{Reptile} \cite{REPTIL} & $0.992$ \\
		\textit{Finstreder (Quartznet)} & $0.992$ \\
		\textit{Saxon et al.} \cite{EEGVA} & $0.994$ \\
		\textit{AT-AT} \cite{ATAT} & $0.995$ \\
		\textit{Finstreder (Conformer)} & $0.995$ \\
		\textit{Borgholt et al.} \cite{DNASR} & $0.996$ \\
		\textit{Seo et al.} \cite{PTNCTI} & $0.997$ \\
		\textit{Qian et al.} \cite{SPLAPT} & $0.997$ \\
		\textit{Kim et al.} \cite{TKNDIS} & $0.997$ \\
		\textit{Finstreder (Quartznet) + AMT} & $0.997$ \\
		\bottomrule
	\end{tabular}
\end{table}

\subsection{Textual inputs}

Instead of decoding CTC-labels which were predicted from an audio input, it is also possible to use the generated LG-FSTs for NLU extraction from textual inputs.
A very simple approach, which was tested here, is to convert the textual input to CTC-labels. Table~\ref{tab:nluo} shows that NLU parsing with \textit{finstreder} can outperform traditional approaches on simple datasets (in \textit{SmartSpeaker} only the artist needs to be extracted), but falls behind if the datasets are more complex (see next chapter for explanation).

\comment{
\begin{table}[!htbp]
	\footnotesize
	\caption{NLU only test with textual input on \textit{Smart\-Speaker}~(above) and \textit{SmartLights} (below).}
	\label{tab:nluo}
	\centering
	\begin{tabular}{lcc}
		\toprule
		\textbf{} & \textbf{Accuracy} & \textbf{WER} \\
		\midrule
		\textit{Jaco (Rasa)} & $0.977$ & $-$ \\
		\textit{Finstreder} & $0.994$ & $0.153$ \\
		\midrule
		\textit{Jaco (Rasa)} & $0.960$ & $-$ \\
		\textit{Finstreder} & $0.889$ & $0.054$ \\
		\bottomrule
	\end{tabular}
\end{table}
}

\comment{
\begin{table}[!htbp]
	\footnotesize
	\caption{NLU only test with textual inputs.}
	\vspace{-5pt}
	\label{tab:nluo}
	\centering
	\begin{tabular}{lcccc}
		\toprule
		\textbf{} & \textbf{SmartSpeaker} & \textbf{WER} & \textbf{SmartLights} & \textbf{WER} \\
		\midrule
		\textit{Jaco (Rasa)} & $0.977$ & $-$ & $0.960$ & $-$ \\
		\midrule
		\textit{Jaco (Rasa)} & $0.977$ & $-$ & $0.960$ & $-$ \\
		\textit{Finstreder} & $0.994$ & $0.153$ & $0.889$ & $0.054$ \\
		\bottomrule
	\end{tabular}
\end{table}
}

\begin{table}[!htbp]
	\footnotesize
	\caption{NLU only test with textual inputs.}
	\vspace{-5pt}
	\label{tab:nluo}
	\centering
	\begin{tabular}{lcccc}
		\toprule
		\textbf{} & \multicolumn{2}{c}{\textbf{SmartSpeaker}} & \multicolumn{2}{c}{\textbf{SmartLights}} \\
		\textbf{} & \textbf{Accuracy} & \textbf{WER} & \textbf{Accuracy} & \textbf{WER} \\
		\midrule
		\textit{Jaco (Rasa)} & $0.977$ & $-$ & $0.960$ & $-$ \\
		\textit{Finstreder} & $0.994$ & $0.153$ & $0.889$ & $0.054$ \\
		\bottomrule
	\end{tabular}
\end{table}

Instead of assigning a probability of $1$ to the actual character and $0$ to the rest, a very high probability around $0.99$ is used and the other characters get a very low probability around $0.001$. The smaller value includes some random noise, which is important for the \textit{top-k} optimization, to ensure that not always the same characters are chosen. Another important note is that between every character a timestep containing the \textit{blank} symbol as highest probability is added. This has two reasons. First, it ensures that repeated characters are not merged into one, and second, it greatly improved the accuracy in some experiments. An explanation might be that it allows the model to slightly change words or invent new ones, if the input sentence doesn't match the training examples very well.

\subsection{Limitations}

The FST-based approach of \textit{finstreder} also has some limitations which should be mentioned. First, it does not work with open questions or commands and can only recognize predefined lookup values. The performance also decreases if there is a large difference between the textual training examples and the test sentences. These limitations can be seen in the \textit{Spoken Language Understanding Resource Package (SLURP)} benchmark~\cite{SLURP}, which is currently the largest and most complicated SLU benchmark and includes multiple different domains and open questions like \textit{``give me the weather forecast for this week"}, \textit{``who won the presidential election this year"} or \textit{``if you had to kill someone to save three people would you do it and if so why"}. The results in Table~\ref{tab:slurp} show that \textit{finstreder} only understands less than a half of the questions, which is much less than the baseline presented with the benchmark. The model from \textit{SLURP} uses state-of-the-art STT and NLU models which were finetuned on this dataset.
Testing \textit{Alexa} was planned as well, but was not possible, because the total number of intents and slots included in the dataset was too high and raised an error message when trying to build the skill.

\begin{table}[!htbp]
	\footnotesize
	\caption{Results on \textit{SLURP} dataset.}
	\vspace{-5pt}
	\label{tab:slurp}
	\centering
	\begin{tabular}{p{2.65cm}>{\centering\arraybackslash}p{1.4cm}>{\centering\arraybackslash}p{1.2cm}>{\centering\arraybackslash}p{1.0cm}}
		\toprule
		\textbf{} & \textbf{ScenAct-F1} & \textbf{Entity-F1} & \textbf{SLU-F1} \\
		\midrule
		\textit{Multi-SLURP} \cite{SLURP} & $0.783$ & $0.642$  & $0.708$ \\
		\textit{Finstreder (Quartznet)} & $0.432$ & $0.313$ & $0.380$ \\
		\textit{Finstreder \mbox{(Conformer)}} & $0.531$ & $0.395$ & $0.452$ \\
		\bottomrule
	\end{tabular}
\end{table}

A second limitation is that the decoding process slows down with growing datasets. Small benchmarks like \textit{Barrista} run about $8/4\times$ (\textit{Quartznet}$/$\textit{Conformer}) faster than real-time on a standard desktop CPU (AMD-3700X) using the \textit{tflite}-runtime, but the large \textit{SLURP} run is only $1.6/1.7\times$ faster, and uses a lower \textit{top-k} decoding parameter. The decoding speed is influenced by two characteristics of the STT model: The computation complexity of the model itself, which is much higher for the \textit{Conformer} model compared to the \textit{Quartznet} model, as well as the number of CTC-timesteps in the model outputs, where the \textit{Conformer} has half the length of the \textit{Quartznet} model, which has a growing impact on larger FST models.
For future development, two options could be interesting to improve the decoding speed for large models: Testing decoders from the \textit{Kaldi} project \cite{KALDI}, which are optimized on large FSTs, and splitting up the skills into sub-grammars, similar to the approach of \textit{Alexa}, which requires an activation phrase to start a skill after speaking the wake-word. This could be implemented with a small FST, which differentiates between the skill activation word, and enables to select only the Intent-FSTs required for this specific skill for the following decoding step. Such splitting should also have a positive effect on recognition accuracy.

\section{Discussion and Conclusion}
\label{sec:conclu}

In this paper a simple method for direct \textit{Spoken Language Understanding} without training is presented under the name of \textit{finstreder}. As basis \textit{Finite State Transducers} are used and optimized for the task of intent and entity extraction. In multiple benchmarks a performance greater than multiple direct SLU or two-step STT+NLU approaches could be achieved.

\vspace{9pt}
The main advantage of \textit{finstreder} over other approaches is that no extra training of the SLU model is required. Only text files describing possible requests are needed, which are easy to create or adjust, and could be distributed, similar as in the voice assistants \textit{Jaco}, \textit{Alexa} or others, through shareable skills.

Building the proposed SLU-FST models for the investigated benchmarks took between 1 second (\textit{FluentSpeechCommands, Barrista, SmartLights}) and about 30 seconds (\textit{SLURP}) on a standard desktop computer (AMD-3700X). 
Compared to \textit{Jaco}~\cite{JACO} this is much faster, training the NLU models of \textit{Rasa} on the same hardware took about \mbox{15 minutes} for \textit{Barrista} and about \mbox{20-25 minutes} for \textit{SmartLights} and \textit{SmartSpeaker}, even though the hyperparameters were optimized for training speed.

The models can also be built and used on edge-devices like a RaspberryPi\,4. Building the model for \textit{FluentSpeechCommands}, in which the request are most similar to a simple voice assistant, took \mbox{$2.2$ seconds} (compared to \mbox{$0.7$ seconds} on the desktop computer). The benchmark itself then runs about~$1.1/1.5\times$ (\textit{Quartznet}$/$\textit{Conformer}) faster than real time using the quantized \textit{tflite} models. As mentioned in the last chapter, it can be seen here that on the RaspberryPi, the reduction of the number of CTC-steps outweighs the slower inference speed of the \textit{Conformer} model.

Most other approaches did not mention training times or the used hardware, except for the \textit{SpeechBrain} run on \textit{Timers\-AndSuch}~\cite{TASSB}, where published logs indicate that the training took about \mbox{2-3 hours} on unknown hardware, as well as the approach of \textit{Kim et al.} on \textit{FluentSpeechCommands}~\cite{TKNDIS}, which required $8\times$ Nvidia-V100 GPUs to train its network. 

\vspace{9pt}
The presented method is especially interesting for use-cases that have frequent domain changes, relatively small datasets or restricted training possibilities. For example this could be the case in customizable smart home assistants running on edge-devices or on smartphones, where using a cloud service is undesirable, either to be independent of unstable internet connections or due to privacy concerns.


\bibliographystyle{IEEEtran}
\bibliography{mybib}

\end{document}